\documentclass[11pt]{article}

\usepackage{acl}

\usepackage{times}
\usepackage{latexsym}

\usepackage[T1]{fontenc}
\usepackage[utf8]{inputenc}

\usepackage{silence}
\usepackage{microtype}

\usepackage{inconsolata}

\usepackage{graphicx}

\usepackage{arydshln}

\usepackage{amsmath}
\usepackage{amssymb}
\usepackage{pifont}% http://ctan.org/pkg/pifont

\usepackage{subcaption}

\usepackage{tcolorbox}
\usepackage{multirow}

\usepackage{float}

\usetikzlibrary{shapes.geometric, arrows, positioning}
\tikzset{
  model/.style = {rectangle, draw, minimum width=2.2cm, minimum height=0.9cm, font=\sffamily\tiny},
  question/.style = {rectangle, align=left, font=\sffamily\tiny, text width=6.5cm},
  arrow/.style = {->, thick, shorten <=1pt, shorten >=1pt},
  iterlabel/.style = {font=\sffamily\tiny\bfseries, anchor=north east, inner sep=1pt}
}

\usepackage{pifont}% http://ctan.org/pkg/pifont

\usepackage{lipsum}

\newcommand\blfootnote[1]{%
  \begingroup
  \renewcommand\thefootnote{}\footnote{#1}%
  \addtocounter{footnote}{-1}%
  \endgroup
}

\title{How DDAIR you?\\Disambiguated Data Augmentation for Intent Recognition}

\author{Galo Castillo-L\'{o}pez \quad Alexis Lombard \quad Nasredine Semmar \quad Ga\"{e}l de Chalendar \\
        Universit\'{e} Paris-Saclay, CEA, List, Palaiseau, France \\ \small{\{\texttt{galo-daniel.castillolopez, alexis.lombard, nasredine.semmar, gael.de-chalendar\}\texttt{@cea.fr}  }}}

\begin{document}
\maketitle
\begin{abstract}
Large Language Models (LLMs) are effective for data augmentation in classification tasks like intent detection. In some cases, they inadvertently produce examples that are ambiguous with regard to untargeted classes. We present DDAIR (\textbf{D}isambiguated \textbf{D}ata \textbf{A}ugmentation for \textbf{I}ntent \textbf{R}ecognition) to mitigate this problem. We use Sentence Transformers to detect ambiguous class-guided augmented examples generated by LLMs for intent recognition in low-resource scenarios. We identify synthetic examples that are semantically more similar to another intent than to their target one. We also provide an iterative re-generation method to mitigate such ambiguities. Our findings show that sentence embeddings effectively help to (re)generate less ambiguous examples, and suggest promising potential to improve classification performance in scenarios where intents are loosely or broadly defined.

\end{abstract}

\section{Introduction}
\label{sec:introduction}
\blfootnote{This paper has been accepted for publication at EACL 2026 and corresponds to the author's version of the work.}
Large Language Models have been proposed and effectively used for data augmentation in several NLP tasks in low-resource scenarios, including intent recognition \cite{wang2024lambda,si2025unified,meguellati2025llm}. A common approach to generate synthetic data from LLMs is through multiple LLM calls, where a single example is generated per call, or a set of examples belonging to a target class are generated in the same inference. While such a practice can produce useful synthetic utterances for intent recognition \cite{benayas2024enhancing}, the lack of information about other intents in the generative process may cause ambiguous\footnotemark outputs. For instance, an augmented utterance that is generated to represent a target intent, could be semantically more similar to another class. Arguably, such ambiguities negatively impact classification systems \cite{liu-etal-2023-afraid}. \footnotetext{In this work, our definition of ``ambiguous'' is task-related rather than linguistic.}

\begin{figure}[t]
\centering
\includegraphics[width=1.00\linewidth]{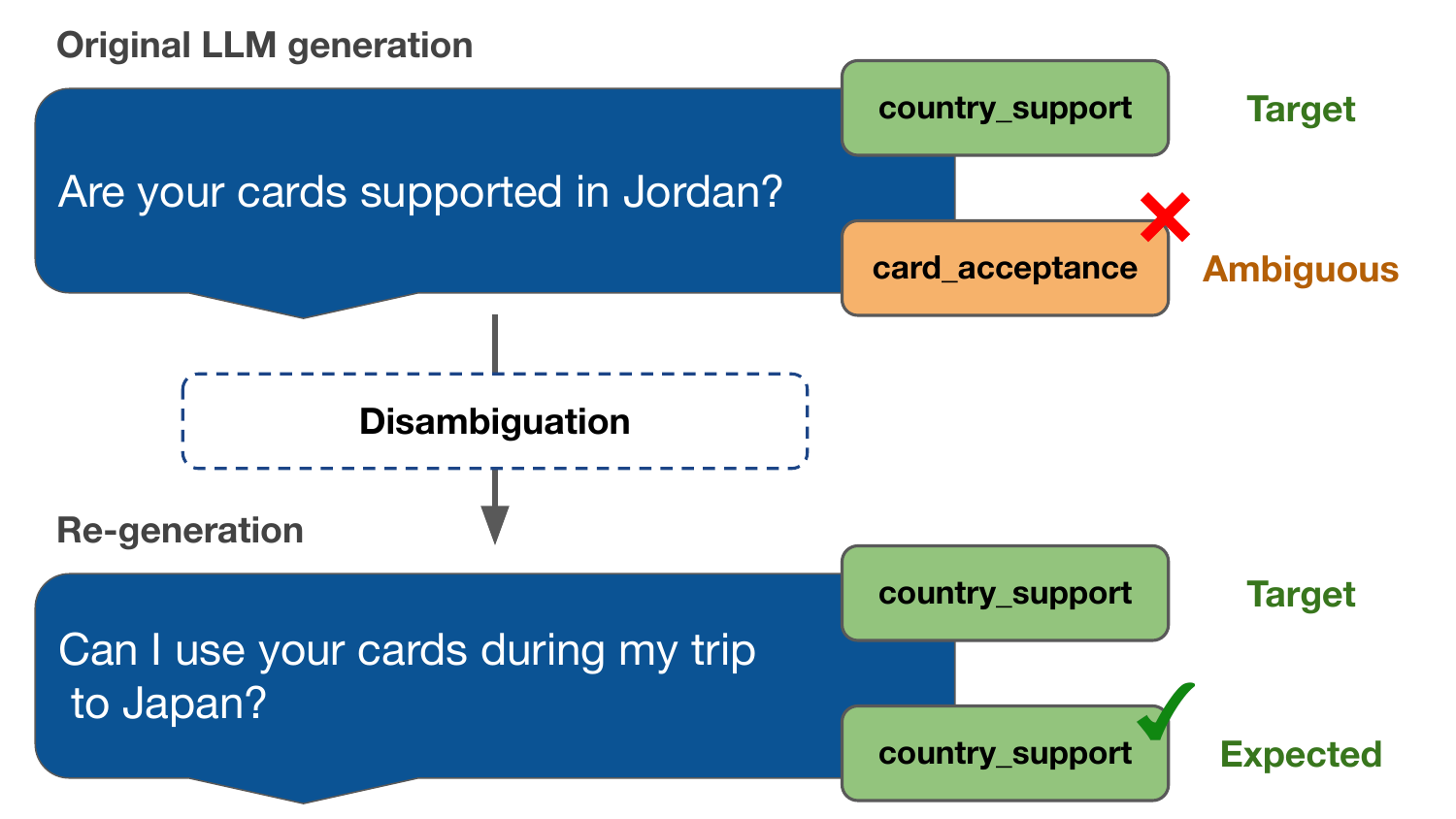}
\caption[]{\textbf{Top:} A LLM-generated utterance for the target label \texttt{country support} from the banking domain\footnotemark. The synthetic utterance is semantically similar to the \texttt{card acceptance} intent. \textbf{Bottom:} After one disambiguation iteration, the re-generated utterance shows no ambiguity with respect to any class from the label space.}
\label{fig:ambiguous_utt_wrt_intents}
\end{figure}

In this paper, we study the impact of ambiguous augmented examples from LLMs for intent recognition. We propose re-generating detected ambiguous utterances to improve augmented data quality. Our method, named DDAIR, consists of combining sentence transformers and LLMs to detect and re-generate ambiguous synthetic examples. We first employ sentence transformers to automatically detect ambiguous LLM-generated utterances, by semantically comparing synthetic utterances with few-shot examples from the training set (Figure~\ref{fig:ambiguous_utt_wrt_intents} upper part). Then, detected ambiguous examples are re-generated using the LLM, including in-context learning examples, the target intent and the ambiguous (unexpected) intent names in the prompt (Figure~\ref{fig:ambiguous_utt_wrt_intents} lower part). Our approach is iterative, such that the detect-and-generate process can be repeated as long as the augmented utterance remains ambiguous.\footnotetext{Label examples from the BANKING77 corpus.}

The main goal of our work is to investigate how detecting and treating ambiguous utterances impacts systems which are fine-tuned with examples generated by LLMs. We assess our method on models from two LLM families, and three sentence encoders. Our results show that sentence transformers can be effectively used to detect ambiguous generations, and in combination with LLMs, reduce the ambiguity of augmented data.

\section{Related Work}
\label{sec:related_work}
Recent work has made key progress on intent detection \cite{atuhurra2024domain}. LLMs have been widely used as intent classifiers \cite{arora-etal-2024-intent,park-etal-2025-dynamic,castillo-lopez-etal-2025-intent}, whereas their application for data augmentation in low-resource scenarios remains relatively underexplored \cite{lin-etal-2024-generate,beatricia2024text}. Previous studies have found that LLMs consistently perform better than traditional methods at augmenting data for intent recognition \cite{benayas2024enhancing}. \citet{sahu-etal-2022-data} observed that the examples generated by LLMs are less helpful in tasks with semantically close intents, as models tend to generate utterances that belong to a semantically related intent to the target class.

Generating and refining synthetic examples for intent detection was studied in \cite{lin-etal-2024-generate}. Their two-stage approach involves using LLMs to generate utterances, followed by fine-tuning a smaller sequence-to-sequence model in a full-shot setting for unseen intents. In contrast, our work assumes limited data scenarios and does not rely on full-shot settings. \citet{lin-etal-2023-selective} propose using Pointwise $\mathcal{V}$-Information \cite{pmlr-v162-ethayarajh22a} to identify and retain the most informative synthetic utterances for training, leading to improved classification performance. While their approach filters out low-quality augmented data, our method focuses on enhancing it.

\section{Experimental Procedure}
\label{sec:experiments}
We propose a data augmentation method for intent recognition in few-shot scenarios. The intent recognition task involves classifying a user utterance $u$ into an intent label $y \in \mathcal{Y}$. To address data scarcity, we leverage large language models to generate synthetic utterances conditioned on intent names to form an augmented dataset. We propose the use of sentence transformers to encode both the synthetic utterances and the in-context learning examples to detect ambiguous (i.e., problematic) utterances and provide a disambiguation strategy.

\subsection{Utterance Generation with LLMs}
\label{sec:utterance_generation}
We employ an in-context learning approach (ICL) to generate synthetic utterances on the instruct versions of two LLMs: Mistral 7B and Llama-3 8B. Our experiments consider from 2 to 5 randomly selected (without replacement) ICL examples per intent class from the training set to simulate a $n$-shot scenario. To reduce the variability introduced by random sampling, we carry out 5 independent sampling rounds for each ICL setting. This results in 5 distinct sets of ICL examples for each $n$-shot setup, where $n \in \{2, 3, 4, 5\}$, enabling a more robust evaluation. We perform one LLM call for each generated utterance. Generation prompts include the task instruction, ICL examples and intent name. More information about our use of LLMs is described in Appendix \ref{sec:app_llms}.

\subsection{Ambiguous Utterance Detection}
\label{sec:ambiguous_detection}
In this work, we propose the detection and treatment of ambiguous generations to improve the quality of synthetic data. We use sentence transformers to encode the generated utterances and ICL examples, to detect ambiguous examples. We consider that a generated utterance is ambiguous if it lies far from the $n$ ICL examples of the target intent (in some embedding space) and closer to the examples of an untargeted intent. We hypothesize that such ambiguous examples can damage the performance of intent recognition models. In particular, let $f(\cdot)$ denote the sentence encoder and $f(u)$ represent the embedding of an utterance $u$. For every utterance $u_i$ in the training set, we obtain its representation $f(u_i)$. For each intent label \( y \in \mathcal{Y} = \{1, \dots, m\} \), we compute a class-specific centroid vector \( c_y \) by averaging the embeddings of its corresponding ICL examples. Let \( \{u_1^{(y)}, \dots, u_n^{(y)}\} \) denote the set of \( n \) ICL utterances labeled with intent \( y \), then the centroid for class \( y \) is then computed as 
%\[
$c_y = \frac{1}{n} \sum_{i=1}^{n} f(u_i^{(y)})$.
%\]
This results in one centroid vector $ c_y \in \mathbb{R}^d$ for each intent class $y \in \mathcal{Y}$, where $d$ is the sentence embedding dimensionality. We define a synthetic utterance $u_j$ as ambiguous if $\hat{y}_j \ne y_j$, where $y_j \in \mathcal{Y}$ is its target intent and $\hat{y}_j \in \mathcal{Y}$ is its neareast centroid's class, i.e., the label of the nearest centroid does not match its target label. We use BGE, MPNet and MiniLM-L6 to compute sentence embeddings. See \ref{sec:app_sentence_transformers} for more details about the sentence transformers.

\subsection{Synthetic Utterance Disambiguation}
\label{sec:disambiguation}
We follow two strategies to treat ambiguous generations. First, similarly as in \cite{lin-etal-2023-selective}, we drop all ambiguous generated examples produced by the LLMs. Hence, we only keep synthetic utterances that lie in the neighborhood of the target intent ICL examples. The second approach consists in a multi-step strategy of detection and re-generations. We detect ambiguous generated utterances, as described in Section \ref{sec:ambiguous_detection}, and use a disambiguation --i.e., re-generation-- prompt (see \ref{sec:app_llms}). Our aim is to semantically align the synthetic utterances to the correct intent label. We conduct the disambiguation procedure iteratively up to 3 times in this work.

\subsection{Intent Classification}
\label{sec:classification}
To asses the utility of our proposed data augmentation approach, we fine-tune uncased BERT\textsubscript{BASE} on the classification task \cite{devlin2019bert}. Models are fine-tuned using both the few-shot and generated utterances. We also conduct a set of evaluations on baseline experiments using the original LLM generations --without any disambiguation strategy-- with varying numbers of ICL examples and generated utterances, which can be found in Appendix \ref{sec:app_baseline}. See Appendix \ref{sec:app_bert} for more details about the fine-tuning process.

\subsection{Datasets}
\label{sec:datasets}
Three intent detection corpora are used in this work: BANKING77 \cite{casanueva-etal-2020-efficient}, CLINC150 \cite{larson-etal-2019-evaluation} and the MPGT corpus \cite{addlesee-etal-2023-multi}. The BANKING77 corpus consists of 77 intents from the banking domain. CLINC150 comprises 150 intents from 10 domains, such as travel, work, and others. In addition, we use MPGT, a collection of multi-party dialogues between users and a hospital receptionist robot. The original MPGT corpus considers that utterances may have more than one label. Hence, we use the multi-class version of the corpus, proposed in \cite{castillo-lopez-etal-2025-intent}, which comprises 8 intents. More information about the datasets in Appendix \ref{sec:dataset_info}.

\subsection{Evaluation}
\label{sec:evaluation}
The evaluation of our work is two-fold. We first evaluate how our iterative disambiguation approach impacts the utterances' ambiguity. To do so, we compute the ambiguity ratio as the proportion of ambiguous utterances from all the LLM-generated examples after each iteration. Second, we apply Silhouette coefficient analysis, a widely used clustering validation metric balancing intra-cluster cohesion and inter-cluster separation \cite{rousseeuw1987silhouettes}. The metric is regarded as a robust indicator of cluster quality \cite{vendramin2010relative, arbelaitz2013extensive} and has recently been used in intent discovery to assess embedding-space cluster structure \cite{liu2021open, de2023idas, hongDialInLLMHumanAligned2025a, ferrera2025linguistics}

Since clusters are given by the known intent labels, we do not perform clustering ourselves. Instead, we apply Silhouette analysis to assess whether the generated utterances align closely with the ICL examples of their expected intents or drift toward neighboring intent regions. For a given utterance \( i \), let \( a(i) \) be the mean distance between \( i \) and all other utterances within the same cluster (i.e., same intent), and let \( b(i) \) be the minimum mean distance between \( i \) and all utterances in any other cluster (i.e., the nearest neighboring intent cluster). The Silhouette Coefficient \( s(i) \) is defined as:

\[
s(i) = \frac{b(i) - a(i)}{\max\{a(i), b(i)\}}
\]

The coefficient ranges from \(-1\) to \(1\), where values close to \(1\) indicate that an utterance is well matched to its assigned cluster, values near \(0\) suggest a boundary case, and negative values imply potential misassignment.

Finally, we evaluate our fine-tuned BERT intent classifiers using the average macro-F1 over 3 runs with different seeds. As described in \S\ref{sec:utterance_generation}, we sample 5 random n-shot sets, so each reported score averages 15 results. We also implement and evaluate the data-augmentation method from \cite{lin-etal-2023-selective} under our settings for comparison.

\section{Results}
\label{sec:results}
Sections \ref{sec:res_disambiguation} and \ref{sec:res_silhouette} provide two different analyses to assess our disambiguation strategy to reduce ambiguity on augmented utterances. We evaluate the utility of our disambiguation strategy to produce higher quality data for the intent classification task in Section \ref{sec:res_classification}.

\begin{figure*}[t]
    \centering
    \begin{subfigure}{0.495\linewidth}
        \centering
        \includegraphics[width=1.0\linewidth]{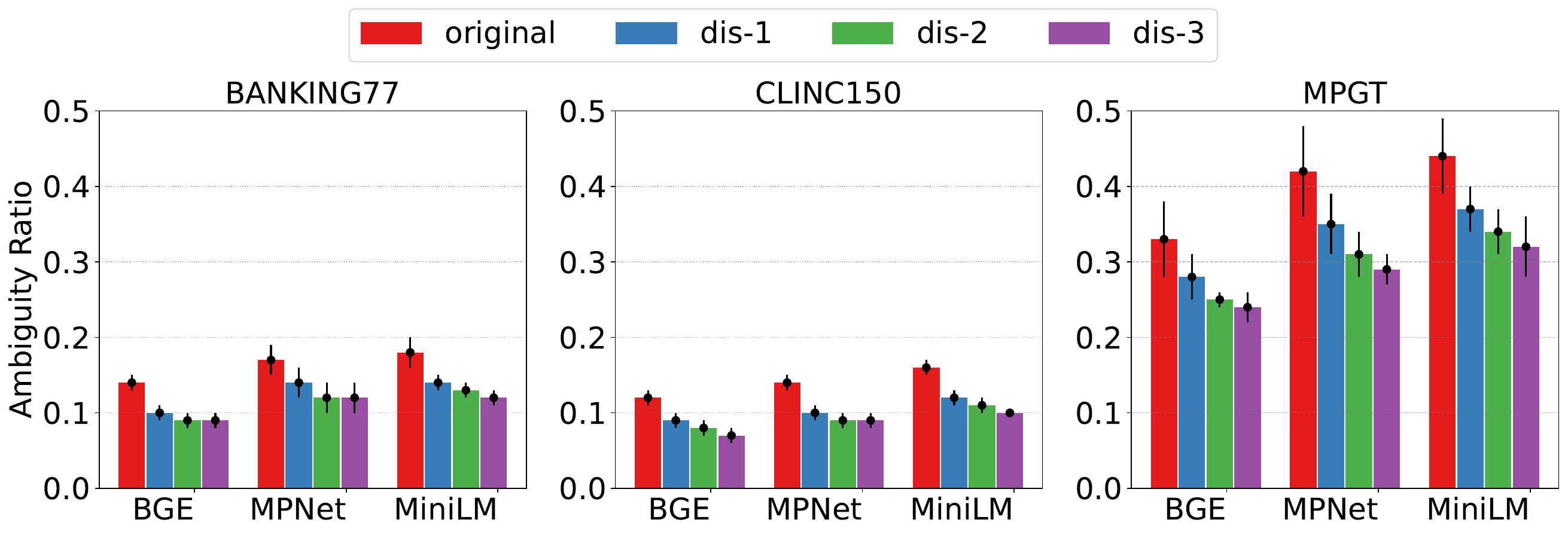}
        \caption{Mistral 7B.}
    \end{subfigure}
    \begin{subfigure}{0.495\linewidth}
        \centering
        \includegraphics[width=1.0\linewidth]{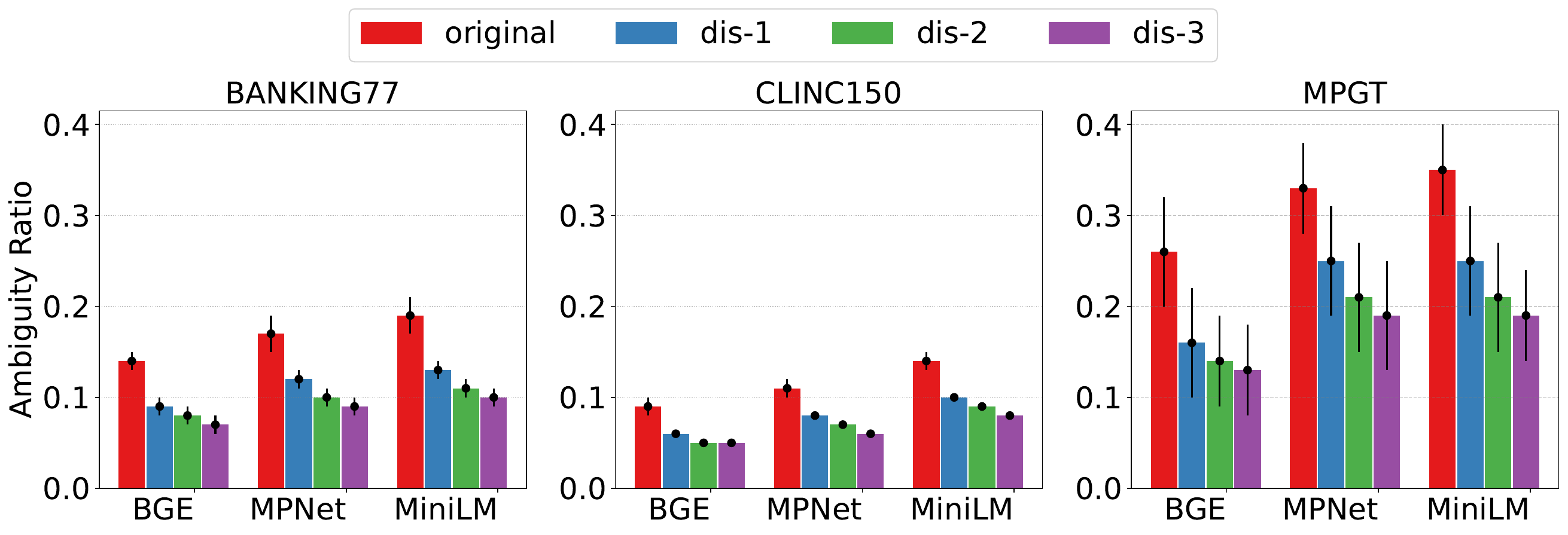}
        \caption{Llama-3 8B.}
    \end{subfigure}
    \caption{Ambiguity ratios of the original generations and re-generations after multiple iterative disambiguation steps (dis-1, dis-2, dis-3). A lower ratio indicates a lower proportion of ambiguous generated utterances. In all scenarios, ambiguity ratios decrease after disambiguation steps. MPGT is the most ambiguous corpus.}
    \label{fig:ambiguity_ratios}
\end{figure*}

\begin{table*}[!t]
\tiny
\centering
% $\heartsuit\varheart\diamondsuit\vardiamond\clubsuit\spadesuit$
\begin{tabular}{c l | c c c | c c c | c c c}
% \begin{tabular}{l l c c c : c c c}
 \hline
  & Dis. & \multicolumn{3}{c|}{BANKING77} & \multicolumn{3}{c}{CLINC150} & \multicolumn{3}{c}{MPGT}\\
\cline{3-11}

$n$-shot & Strategy & BGE & MPNet & MiniLM & BGE & MPNet & MiniLM & BGE & MPNet & MiniLM\\ 

\hline
% \textbf{Mistral 7B} &  &   &  &  &  &  &  &  &  & \\
\multicolumn{2}{l}{\textbf{Mistral 7B}}  &   &  &  &  &  &  &  &  & \\

2 & w/o dis. & 56.62 $\pm$ .40  & 56.62 $\pm$ .40 & 56.62 $\pm$ .40 & 75.82 $\pm$ .19 & 75.82 $\pm$ .19 & 75.82 $\pm$ .19 & 62.17 $\pm$ 1.92 & 62.17 $\pm$ 1.92 & 62.17 $\pm$ 1.92 \\

% 2 & PVI\footnote{\cite{lin-etal-2023-selective}} & 24.57 $\pm$ .51  & 24.57 $\pm$ .51 & 24.57 $\pm$ .51 & 75.82 $\pm$ .19 & 75.82 $\pm$ .19 & 75.82 $\pm$ .19 & 62.17 $\pm$ 1.92 & 62.17 $\pm$ 1.92 & 62.17 $\pm$ 1.92 \\

2 & PVI (\textbf{*}) & 32.99 $\pm$ .49  & 32.99 $\pm$ .49 & 32.99 $\pm$ .49 & 46.11 $\pm$ .30 & 46.11 $\pm$ .30 & 46.11 $\pm$ .30 & 56.22 $\pm$ 1.52 & 56.22 $\pm$ 1.52 & 56.22 $\pm$ 1.52 \\

2 & Drop & 51.12 $\pm$ .33 & 49.58 $\pm$ .44 & 48.42 $\pm$ .52 & 67.90 $\pm$ .37 & 67.33 $\pm$ .27 & 66.11 $\pm$ .16 & 68.00 $\pm$ 1.18 & \underline{75.15 $\pm$ 1.07} & \textbf{69.75 $\pm$ 1.72} \\

2 & Dis-1 & \textbf{57.43 $\pm$ .38} & \textbf{57.03 $\pm$ .30} & \textbf{57.28 $\pm$ .37} & 75.95 $\pm$ .14 & \underline{76.11 $\pm$ .14} & \underline{76.27 $\pm$ .15} & \underline{68.08 $\pm$ 1.14} & 68.19 $\pm$ 1.19 & 69.54 $\pm$ 1.44 \\

2 & Dis-2 & \underline{57.37 $\pm$ .42} & \underline{56.96 $\pm$ .31} & 56.93 $\pm$ .38 & \underline{76.19 $\pm$ .16} & 76.10 $\pm$ .18 & 76.40 $\pm$ .18 & \textbf{68.36 $\pm$ .88} & 70.43 $\pm$ .93 & \underline{69.67 $\pm$ 1.23} \\

2 & Dis-3 & 57.07 $\pm$ .44 & 56.66 $\pm$ .34 & \underline{57.16 $\pm$ .27} & \textbf{76.43 $\pm$ .18} & \textbf{76.32 $\pm$ .13} & \textbf{76.56 $\pm$ .17} & 65.81 $\pm$ 1.45 & \textbf{72.05 $\pm$ 1.07} & 68.98 $\pm$ 1.21 \\

% \hdashline 
% \hdashline[0.5pt/5pt]
\hdashline[1.0pt/1pt]
% \hline

5 & w/o dis. & 71.14 $\pm$ .17  & 71.14 $\pm$ .17 & 71.14 $\pm$ .17 & 85.11 $\pm$ .11 & 85.11 $\pm$ .11 & 85.11 $\pm$ .11 & 75.69 $\pm$ .49 & 75.69 $\pm$ .49 & 75.69 $\pm$ .49 \\

5 & PVI (\textbf{*}) & 55.33 $\pm$ .37  & 55.33 $\pm$ .37 & 55.33 $\pm$ .37 & 67.72 $\pm$ .43 & 67.72 $\pm$ .43 & 67.72 $\pm$ .43 & 73.08 $\pm$ 0.79 & 73.08 $\pm$ 0.79 & 73.08 $\pm$ 0.79 \\

5 & Drop & 70.45 $\pm$ .19 & 70.24 $\pm$ .23 & 70.51 $\pm$ .21 & 83.27 $\pm$ .14 & 82.29 $\pm$ .18 & 81.85 $\pm$ .13 & \textbf{81.27 $\pm$ .65} & \textbf{79.25 $\pm$ .46} & \textbf{78.38 $\pm$ .79} \\

5 & Dis-1 & 71.53 $\pm$ .23 & \textbf{72.12 $\pm$ .15} & \underline{71.67 $\pm$ .18} & 85.09 $\pm$ .15 & 85.12 $\pm$ .11 & \underline{85.11 $\pm$ .10} & 76.78 $\pm$ .52 & 76.67 $\pm$ .52 & 75.05 $\pm$ .73 \\

5 & Dis-2 & \underline{71.53 $\pm$ .20} & \underline{71.89 $\pm$ .18} & 71.64 $\pm$ .17 & \underline{85.23 $\pm$ .10} & \underline{85.17 $\pm$ .08} & 85.07 $\pm$ .12 & 77.28 $\pm$ .50 & \underline{76.71 $\pm$ .64} & \underline{77.03 $\pm$ .51} \\

5 & Dis-3 & \textbf{71.61 $\pm$ .18} & 71.70 $\pm$ .15 & \textbf{71.67 $\pm$ .16} & \textbf{85.31 $\pm$ .12} & \textbf{85.22 $\pm$ .12} & \textbf{85.16 $\pm$ .13} & \underline{77.28 $\pm$ .50} & 75.40 $\pm$ .48 & 76.03 $\pm$ .83 \\

\hline

% \textbf{Llama-3 8B} &  &   &  &  &  &  &  &  &  & \\
\multicolumn{2}{l}{\textbf{Llama-3 8B}}  &   &  &  &  &  &  &  &  & \\

2 & w/o dis. & 59.30 $\pm$ .26 & 59.30 $\pm$ .26 & 59.30 $\pm$ .26 & 77.19 $\pm$ .24 & 77.19 $\pm$ .24 & 77.19 $\pm$ .24 & 70.95 $\pm$ 1.09 & 70.95 $\pm$ 1.09 & 70.95 $\pm$ 1.09 \\

2 & PVI (\textbf{*}) & 34.02 $\pm$ .54  & 34.02 $\pm$ .54 & 34.02 $\pm$ .54 & 44.45 $\pm$ .48 & 44.45 $\pm$ .48 & 44.45 $\pm$ .48 & 56.45 $\pm$ 1.30 & 56.45 $\pm$ 1.30 & 56.45 $\pm$ 1.30 \\

2 & Drop & 52.90 $\pm$ .42 & 50.74 $\pm$ .34 & 49.53 $\pm$ .44 & 71.29 $\pm$ .28 & 69.75 $\pm$ .23 & 67.25 $\pm$ .22 & \textbf{72.73 $\pm$ .39} & \textbf{72.16 $\pm$ .11} & 71.94 $\pm$ .88 \\

2 & Dis-1 & \underline{60.33 $\pm$ .28} & 59.98 $\pm$ .29 & \textbf{60.84 $\pm$ .31} & 77.67 $\pm$ .17 & \textbf{77.41 $\pm$ .21} & \textbf{78.24 $\pm$ .21} & \underline{71.56 $\pm$ .96} & 71.50 $\pm$ .91 & \textbf{73.04 $\pm$ .89} \\

2 & Dis-2 & 60.31 $\pm$ .16 & \textbf{60.45 $\pm$ .27} & 60.35 $\pm$ .26 & \underline{77.74 $\pm$ .21} & \underline{77.51 $\pm$ .19} & 77.93 $\pm$ .19 & 69.17 $\pm$ 1.05 & 71.45 $\pm$ .81 & 70.86 $\pm$ .30 \\

2 & Dis-3 & \textbf{60.60 $\pm$ .15} & \underline{60.35 $\pm$ .20} & \underline{60.52 $\pm$ .23} & \textbf{77.95 $\pm$ .17} & 77.48 $\pm$ .15 & \underline{78.13 $\pm$ .22} & 70.96 $\pm$ 1.00 & \underline{71.78 $\pm$ .36} & \underline{72.81 $\pm$ .51} \\

% \hdashline
% \hdashline[0.5pt/5pt]
\hdashline[1.0pt/1pt]
% \hline

5 & w/o dis. & 72.14 $\pm$ .19 & 72.14 $\pm$ .18 & 72.14 $\pm$ .18 & 85.26 $\pm$ .13 & 85.26 $\pm$ .13 & 85.26 $\pm$ .13 & 80.26 $\pm$ .47 & 80.26 $\pm$ .47 & 80.26 $\pm$ .47 \\

5 & PVI (\textbf{*}) & 55.12 $\pm$ .41  & 55.12 $\pm$ .41 & 55.12 $\pm$ .41 & 66.54 $\pm$ .36 & 66.54 $\pm$ .36 & 66.54 $\pm$ .36 & 71.53 $\pm$ 0.98 & 71.53 $\pm$ 0.98 & 71.53 $\pm$ 0.98 \\

5 & Drop & 70.38 $\pm$ .16 & 70.15 $\pm$ .16 & 69.65 $\pm$ .22 & 84.13 $\pm$ .16 & 83.81 $\pm$ .15 & 83.01 $\pm$ .14 & \textbf{83.04 $\pm$ .48} & \textbf{80.54 $\pm$ .98} & \textbf{82.76 $\pm$ .42} \\

5 & Dis-1 & 72.57 $\pm$ .18 & 72.30 $\pm$ .17 & \underline{72.50 $\pm$ .17} & \underline{85.36 $\pm$ .13} & 85.32 $\pm$ .15 & 85.32 $\pm$ .12 & 80.02 $\pm$ .47 & \underline{80.36 $\pm$ .56} & 78.83 $\pm$ .52 \\

5 & Dis-2 & \underline{72.64 $\pm$ .16} & \underline{72.54 $\pm$ .13} & \textbf{72.54 $\pm$ .16} & 85.35 $\pm$ .15 & \underline{85.33 $\pm$ .15} & \textbf{85.54 $\pm$ .14} & 79.73 $\pm$ .39 & 79.95 $\pm$ .50 & 79.05 $\pm$ .50 \\

5 & Dis-3 & \textbf{72.70 $\pm$ .12} & \textbf{72.72 $\pm$ .12} & 72.30 $\pm$ .16 & \textbf{85.48 $\pm$ .13} & \textbf{85.41 $\pm$ .14} & \underline{85.50 $\pm$ .15} & \underline{80.50 $\pm$ .47} & 80.27 $\pm$ .51 & \underline{80.70 $\pm$ .28} \\

\hline

\end{tabular}
\caption{Results on the intent classification models, fine-tuned on $n$ few-shot and 10 synthetic examples. Scores correspond to the average macro F1 and their standard deviations over 15 runs. %(5 random sampling processes of $n$ ICL examples and 3 random seeds at fine-tuning). 
Strategies include: not conducting any disambiguation step (w/o dis.); dropping ambiguous generations (Drop); multiple iterative disambiguation steps (Dis-1, Dis-2, Dis-3); and our implementation of the Per-Intent PVI method from \cite{lin-etal-2023-selective} (\textbf{*}). Scores in \textbf{bold} highlight \textbf{the best} performance, and \underline{underlined} scores represent the \underline{second best} performance in the $n$-shot setup, using a given generative model and a sentence transformer for disambiguation. Results show that it is useful to process ambiguities: the two best performances are always obtained after dropping or disambiguation.}

\label{table:results_main_comb_alter}
\end{table*}

\subsection{Ambiguity Ratio}
\label{sec:res_disambiguation}
\begin{figure}[H]
\centering
\includegraphics[width=0.65\linewidth]{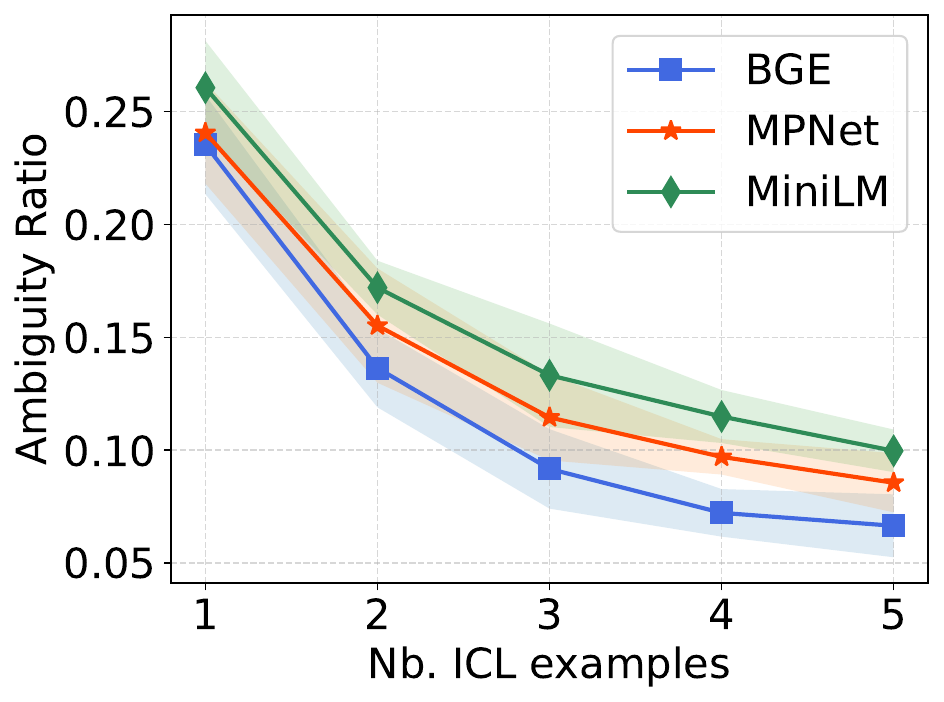}
\caption{Ambiguity ratio by number of in-context learning examples on Mistral 7B generations for the CLINC150 corpus on multiple embedding spaces.}
\label{fig:ambiguity_by_nb_icl}
\end{figure}

Figure~\ref{fig:ambiguity_ratios} displays the ambiguity ratios after each disambiguation step. We observe that ambiguity ratios decrease after every disambiguation step in all scenarios. Our results suggest that we can use sentence transformers alongside LLMs to identify ambiguous synthetic examples and generate less ambiguous utterances for intent recognition. These findings are in line with our Silhouette coefficient analysis in Section~\ref{sec:res_silhouette}. We identify that the MPGT corpus exhibits the highest ambiguity ratios in all settings, presenting ratios about twice as large as the other datasets in similar scenarios. For instance, more than 40\% of the original Mistral generations on MPGT corresponds to ambiguous utterances according to MPNet and MiniLM-L6, while less than 20\% of utterances are labeled as ambiguous on the other datasets. We also note that ambiguity ratios decrease as the number of ICL examples increases in all our settings of LLMs, datasets and embedding spaces. Figure~\ref{fig:ambiguity_by_nb_icl} illustrates such behaviour on the original generations from Mistral 7B for the CLINC150 corpus.

\subsection{Silhouette Coefficient Analysis}
\label{sec:res_silhouette}
\begin{figure}[t]
    \centering
    \includegraphics[width=1.00\linewidth]{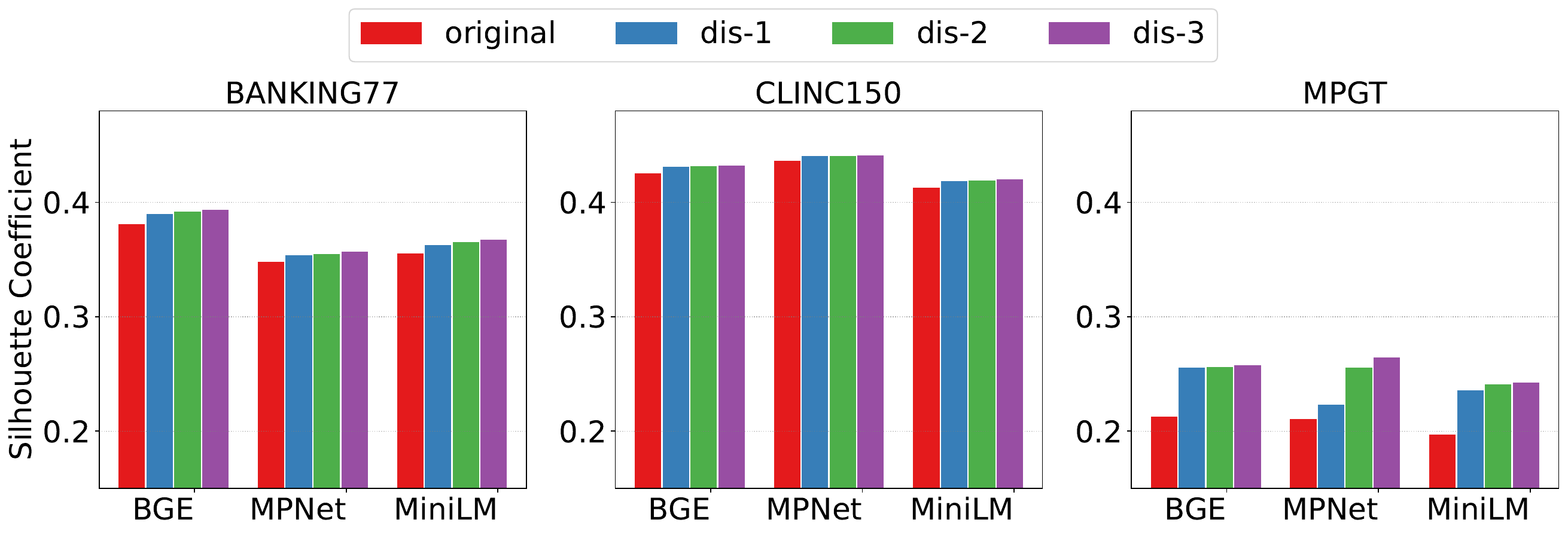}
    \caption{Silhouette coefficients of the original Llama-3 8B generations and re-generations after multiple iterative disambiguation steps (dis-1, dis-2, dis-3). A higher coefficient indicates a better inter-cluster and intra-cluster mean distance relation of the utterance in the embedding space. In all scenarios, the coefficients increase after 3 disambiguation steps.}
    \label{fig:silhouette_analysis}
\end{figure}

Figure~\ref{fig:silhouette_analysis} shows that, under the same experimental settings, disambiguation generally improves cluster cohesion and separation, corroborating our ambiguity ratio findings. On BANKING77 and CLINC150, gains are modest but consistent across encoders. On MPGT, where the baseline Silhouette coefficient is the lowest, improvements are substantial. Corresponding results for Mistral, reported in Figure~\ref{fig:silhouette_analysis_annex} (Appendix~\ref{sec:app_silhouette}), show a first-step regression on MPGT across all encoders, with later steps recovering and surpassing the baseline for MPNet and BGE, while MiniLM-L6 remains close to the baseline. Overall, performing 3 disambiguation steps outperforms the no-disambiguation condition in 17 of 18 settings, indicating the benefit of iterative disambiguation, especially in domains with overlapping intents.

\subsection{Intent Classification}
\label{sec:res_classification}
Our results on the thorough set of experiments without disambiguation, reported on Appendix \ref{sec:app_baseline}, show that better performances are obtained when increasing the numbers of ICL and synthetic examples. Thus, we evaluate our proposed disambiguation strategies considering 2 and 5 ICL examples, as well as 10 augmented examples. Table~\ref{table:results_main_comb_alter} shows the evaluation of the intent recognition models in 2-shot and 5-shot scenarios. We observe that although performance gains are not large when following disambiguation steps in \textbf{5-shot} settings, we consistently obtain improvements on the BANKING77 and CLINC150 datasets after 3 steps for both generators. In most setups (11 out of 12), the best performing models are obtained after 2 or 3 disambiguation iterations in both corpora. On the other hand, our results show that the best strategy for the MPGT corpus is dropping ambiguous generations in all setups, whereas such a strategy is the worst for the other datasets. We obtain improvements on MPGT of up to $\thickapprox$6 and $\thickapprox$3 points on Mistral and Llama generations, respectively. We also find that our methods outperforms our implementation of the PVI-based approach in all setups. We believe that filtering out examples, through the PIV-based approach, results in class imbalance problems at fine-tuning. We provide further analysis on the PVI results in Appendix \ref{sec:app_pvi_experiments}.

Similar to our results in 5-shot, we see that the best scores in all the \textbf{2-shot} experiments are obtained after following one of the disambiguation strategies. For instance, models that are fine-tuned with Mistral generations on MPGT present significant enhancements after dropping or re-generating utterances. Average macro-F1 is increased to up to $\thickapprox$6 points on MPGT. Moreover, we note that larger standard deviations are found on 2-shot experiments than the 5-shot scenario, which might be explained by the instability of the intent centroids given the lower number of examples. We hypothesize that higher performance lifts are seen in the MPGT corpus, compared to the other corpora, due to the high ambiguity on its synthetic examples, as well as to the presence of lower granularity on intent definitions than the other corpora \cite{sahu-etal-2022-data}. This hypothesis is supported by our silhouette coefficient analysis, which shows the largest improvements on MPGT where baseline clustering quality was poorest. We provide further analysis on MPGT ambiguities in Appendix~\ref{sec:app_ambiguity_by_nb_icl_mpgt}. Additionally, qualitative evaluation of our proposed method is provided in Appendix~\ref{sec:qualitative_analysis}.

\section{Conclusions}
\label{sec:conclusions}
We present an iterative data augmentation method to detect and re-generate ambiguous utterances for the intent recognition task. Our work shows that sentence encoders, in combination with LLMs, can be used to reduce the ambiguity ratio of generated utterances and produce examples with better quality for intent detection systems. In line with previous research, we observe that our method significantly improves the performance of systems with loosely or broadly defined intents. We also acknowledge that our proposed disambiguation approach may result in a higher number of LLMs calls. Nevertheless, we argue that such additional computational costs in the generative task might be amortized in the long term. We discuss such a trade-off in Appendix \ref{sec:computational_cost}. We hope that our findings provide insights for future work in data augmentation, not only for intent recognition, but also other text classification tasks.

\section*{Limitations}
\label{sec:limitations}
\paragraph{Ambiguity Definition.} As explained in Section \ref{sec:ambiguous_detection}, we define an utterance as \textit{ambiguous} whether its closest intent centroid, in the selected embedding space, is not its target intent. One important limitation on that definition is the sensibility to outliers of the intent centroid computation. Particularly, there might be cases where certain few-shot examples lie far from the rest of the examples used to compute the centroid. This may cause instability, especially in highly data-scarce scenarios, such as 1-shot or 2-shot settings. We aim to mitigate this issue by computing the element-wise median of the sentence embeddings as an alternative to obtain a more robust estimate of the central tendency for each intent class, notably in the presence of outliers. Let \( \{u_1^{(y)}, \dots, u_n^{(y)}\} \) denote the set of \( n \) ICL utterances labeled with intent \( y \), and let \( f(u_i^{(y)}) \) denote their sentence embeddings. We define the class-specific median vector \( \tilde{c}_y \) as the element-wise median across the embeddings. This results in one median vector \( \tilde{c}_y \in \mathbb{R}^d \) for each intent class \( y \in \mathcal{Y} \), where the median is taken independently over each of the \( d \) embedding dimensions. By analyzing the computed medians, we observed that our results, based on the centroids, were not sensible to outliers. Nonetheless, we believe that future work may consider different strategies to detect ambiguous utterances on the embedding space, such as k-NN, or an adapted k-NN algorithm on the centroids -- the k-Nearest Centroids.

\paragraph{Dependency on Embeddings Quality.} Our proposed ambiguity detection method heavily relies on the sentence embedding space. However, we aim to provide conclusive findings, regardless of the selected embedding space, by conducting experiments over several well-known encoders combined with various generators.

\section*{Ethical Considerations}
Our work essentially consists in refining text generation from Large Language Models for data augmentation. Synthetic utterances generated by LLMs, as other types of LLM-generated texts, may present inherent biases from the models. Hence, outputs from our proposed method rely in controlled generation strategies to mitigate bias and harmful LLM outputs. After a manual analysis of the generated text in our experiments, we do not foresee harm resulting from the experiments conducted in this study. Additionally, our experiments use publicly available corpora, which have been curated prior to our work to prevent malicious actions.

The LLM inference performed in this work to generate synthetic utterances was executed on private infrastructure using a single NVIDIA A100 SXM 40 GB GPU. The infrastructure has a carbon efficiency of 0.432 kgCO$_2$eq/kWh. The total time required for the generative tasks across all experiments was 23.6 hours using Mistral 7B and 32.3 hours using Llama-3 8B. Therefore, the total emissions are estimated at 6.03 kgCO$_2$eq. These estimations are
based on the Machine Learning Impact calculator\footnote{\url{https://mlco2.github.io/impact/}}
\cite{lacoste2019quantifying}.

\section*{Acknowledgments}
We warmly thank our anonymous reviewers for their time and valuable feedback. This publication was made possible by the use of the FactoryIA supercomputer, financially supported by the Ile-de-France Regional Council. This work has been partially funded by the EU project CORTEX$^2$ (under grant agreement: N° 101070192).

% Bibliography entries for the entire Anthology, followed by custom entries
%\bibliography{anthology,custom}
% Custom bibliography entries only
\bibliography{acl_latex}

\appendix
\section{Model Information}
\label{sec:app_model_info}
In this appendix we provide model implementation details of our experiments.

\subsection{BERT}
\label{sec:app_bert}
We use BERT to build intent recognition models. All our runs of BERT fine-tuning were conducted on a single NVIDIA Tesla V100 SXM2 GPU of 32GB. Uncased BERT\textsubscript{BASE} (110M parameters) is used in all the experiments. We used the \texttt{BertForSequenceClassification} class from Hugging Face’s Transformers library \cite{wolf2020transformers} for sequence classification tasks. Table~\ref{table:hyperparameters} shows the hyperparameter configuration we employ. For all intent classification experiments, we report the results averaged over three runs, using random seeds ranging from 0 to 2.
\begin{table}[!ht]
\centering
% \footnotesize
\begin{tabular}{lrr} \hline
hyperparameter & value\\ \hline
eval\_monitor & macro F1 \\
train\_batch\_size & 16 \\
eval\_batch\_size & 16 \\
test\_batch\_size & 16 \\
wait\_patience & 3 \\
num\_train\_epochs & 25 \\
warmup\_proportion & 0.1 \\
lr & 1e-5 \\
\hline\end{tabular}
\caption{Set of hyperparameters used on BERT fine-tuning experiments.}
\label{table:hyperparameters}
\end{table}

\subsection{Large Language Models}
\label{sec:app_llms}
We use Large Language Models to generate --and re-generate-- synthetic utterances for data augmentation. Our experiments use instruct versions of Mistral 7B and Llama-3 8B. Specifically, we use \texttt{Mistral-7B-Instruct-v0.2} \cite{jiang2023mistral7b} and \texttt{Meta-Llama-3-8B-Instruct} \cite{grattafiori2024llama}. Figures \ref{fig:prompt_generation} and \ref{fig:prompt_regeneration} show the prompt templates we use on the generation and re-generation experiments, respectively.

\begin{figure}[htb!]
\centering
\begin{tcolorbox}[colback=gray!5!white, colframe=gray!40!black,
                  title=Generation Prompt,
                  fonttitle=\bfseries,
                  sharp corners=south, boxrule=0.4pt, width=0.95\linewidth]

Give me 1 user's utterance indicating the user intent of "\textcolor{blue}{[intent]}" in the domain "\textcolor{blue}{[domain]}". You can use the following utterances as examples:

\textcolor{blue}{- ICL example 1}

\textcolor{blue}{- ICL example 2}

\textcolor{blue}{...}

\textcolor{blue}{- ICL example N} \\

Your response should only be a JSON object with the following structure: \\

\{"utterance": "generated\_utterance"\}

\end{tcolorbox}
\caption{Prompt template used on all LLM generation experiments. Highlighted text in \textcolor{blue}{blue} varies according to the expected intent and in-context learning examples.}
\label{fig:prompt_generation}
\end{figure}

\begin{figure}[htb!]
\centering
\begin{tcolorbox}[colback=gray!5!white, colframe=gray!40!black,
                  title=Re-generation Prompt,
                  fonttitle=\bfseries,
                  sharp corners=south, boxrule=0.4pt, width=0.95\linewidth]

I generated 1 user's utterance indicating the user intent of "\textcolor{blue}{[intent]}" in the domain "\textcolor{blue}{[domain]}" as follows: "\textcolor{blue}{[ambiguous\_utterance]}". However, the utterance is more similar to the intent "\textcolor{blue}{[most\_similar\_intent]}" which may cause ambiguities. Please refine and disambiguate the generated utterance. You can use the following utterances as examples of the expected intent:

\textcolor{blue}{- ICL example 1}

\textcolor{blue}{- ICL example 2}

\textcolor{blue}{...}

\textcolor{blue}{- ICL example N} \\

Your response should only be a JSON object with the following structure: \\

\{"utterance": "generated\_utterance"\}

\end{tcolorbox}
\caption{Prompt template used on all LLM re-generation experiments (disambiguation). Highlighted text in \textcolor{blue}{blue} varies according to the expected intent, the computed most similar intent, the ambiguous synthetic utterance, and in-context learning examples.}
\label{fig:prompt_regeneration}
\end{figure}

\subsection{Sentence Transformers}
\label{sec:app_sentence_transformers}
Our proposed methods consist in detecting ambiguous augmented examples for intent recognition. To do so, we use sentence embeddings to encode both real and synthetic utterances, and capture semantic representations of training examples. Experiments are performed with three sentence transformers: \texttt{BAAI/bge-base-en-v1.5} \cite{xiaoCPackPackedResources2024}, \texttt{all-mpnet-base-v2} \cite{song2020mpnet}, and \texttt{all-MiniLM-L6-v2} \cite{wang2020minilm}. We used the \texttt{SentenceTransformer} class from the SentenceTransformers library \cite{reimers-2019-sentence-bert} to compute utterance embeddings.

\section{Baseline Experiments}
\label{sec:app_baseline}
\begin{figure*}
    \centering
    
    \begin{subfigure}[t]{0.90\linewidth}
    \begin{subfigure}[t]{0.32\linewidth}
        \centering
        \includegraphics[width=1.0\linewidth]{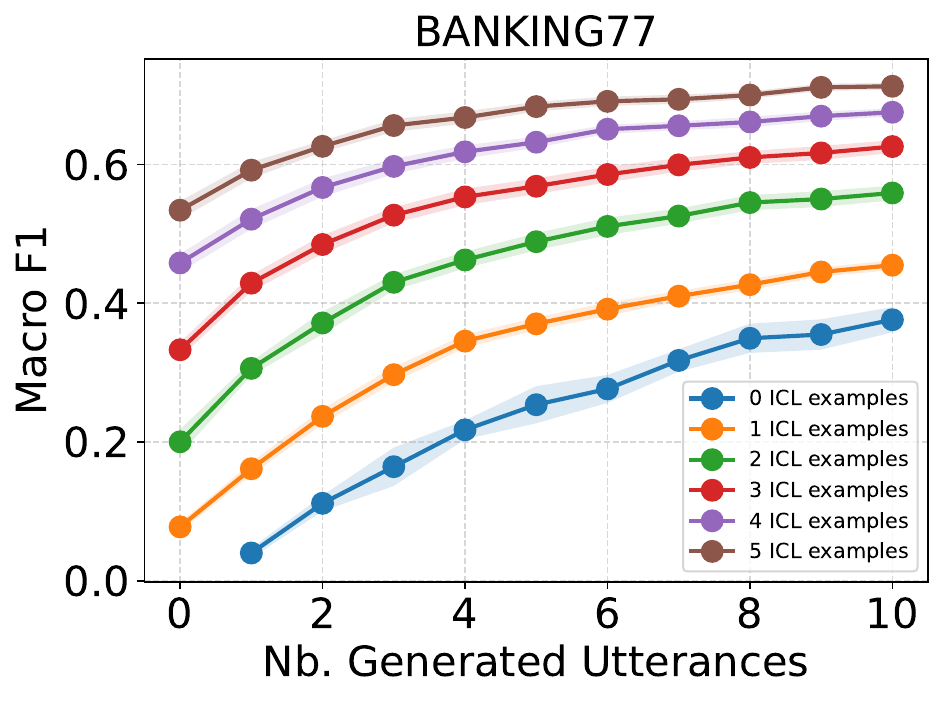}
    \end{subfigure}
    \begin{subfigure}[t]{0.32\linewidth}
        \centering
        \includegraphics[width=1.0\linewidth]{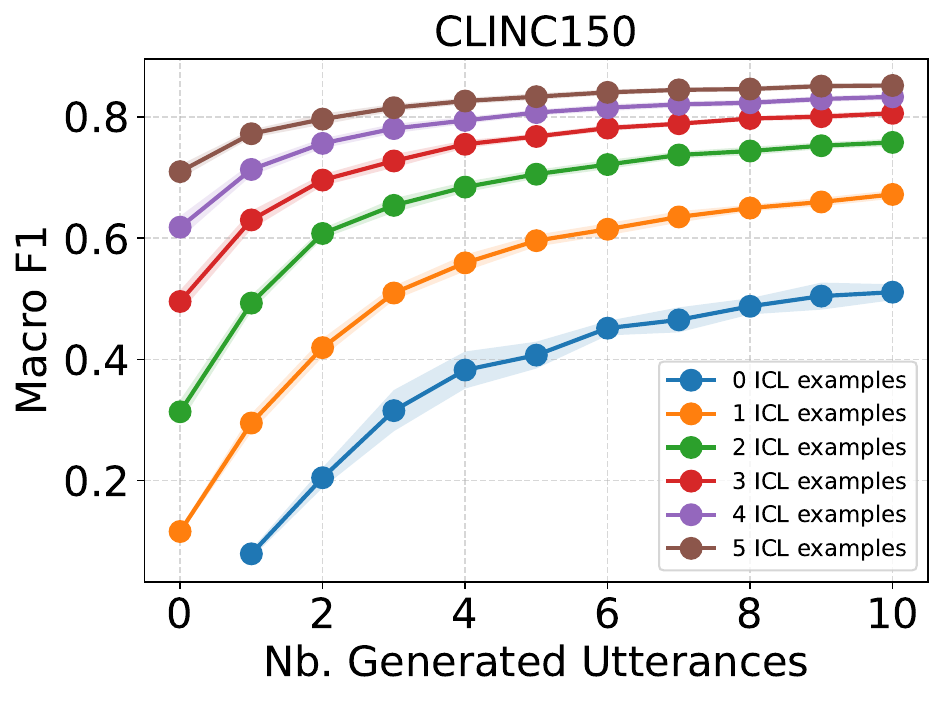}
    \end{subfigure}
    \begin{subfigure}[t]{0.32\linewidth}
        \centering
        \includegraphics[width=1.0\linewidth]{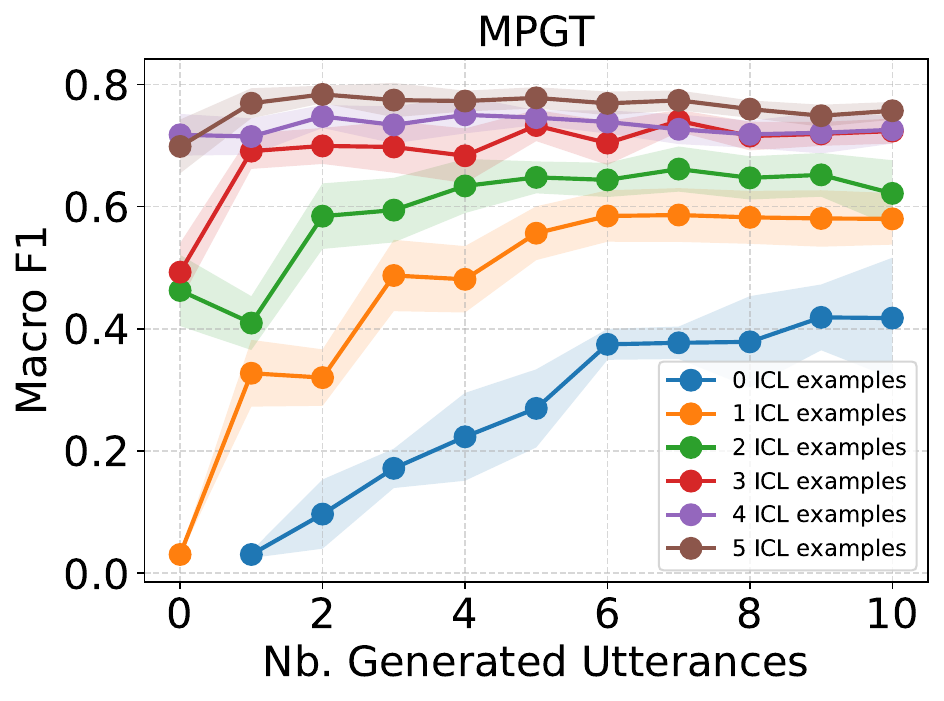}
    \end{subfigure}
    \caption{Mistral 7B.} 
    \end{subfigure}
    \begin{subfigure}[t]{0.90\linewidth}
        \begin{subfigure}[t]{0.32\linewidth}
        \centering
        \includegraphics[width=1.0\linewidth]{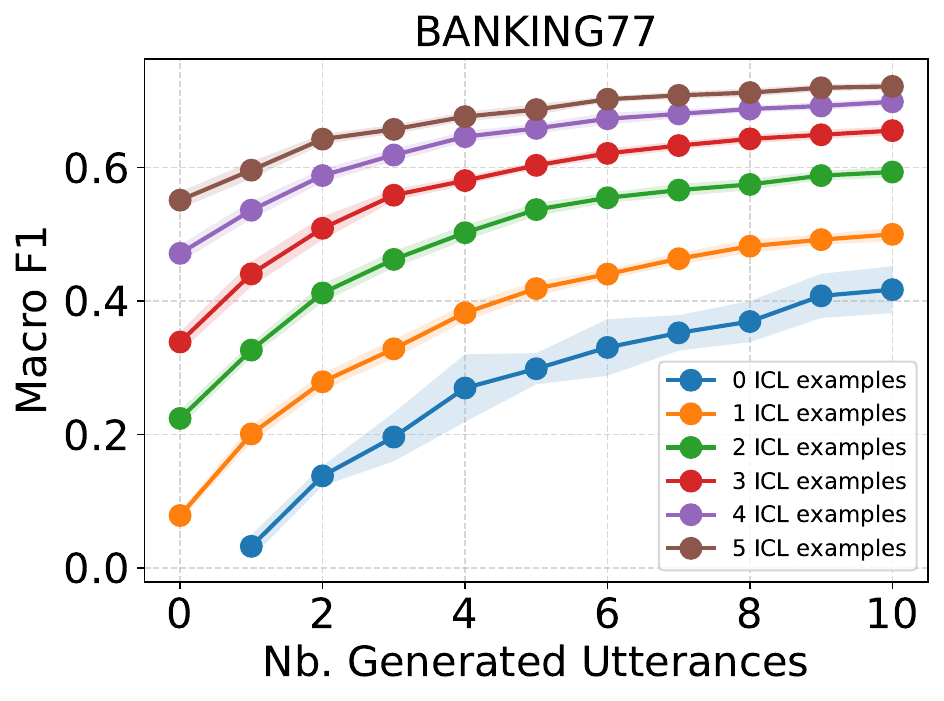}
    \end{subfigure}
    \begin{subfigure}[t]{0.32\linewidth}
        \centering
        \includegraphics[width=1.0\linewidth]{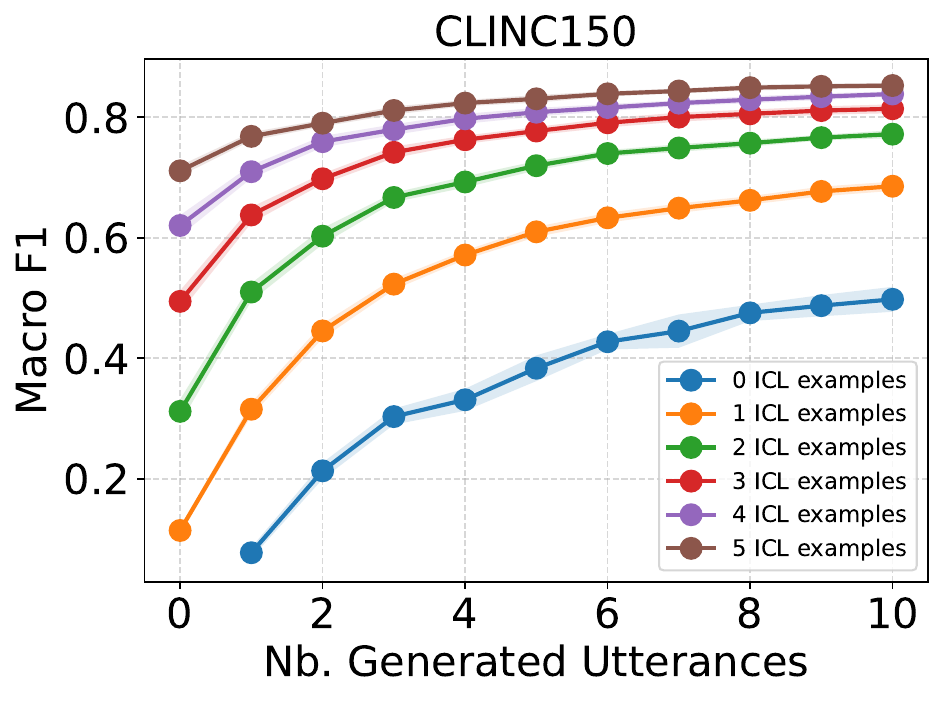}
    \end{subfigure}
    \begin{subfigure}[t]{0.32\linewidth}
        \centering
        \includegraphics[width=1.0\linewidth]{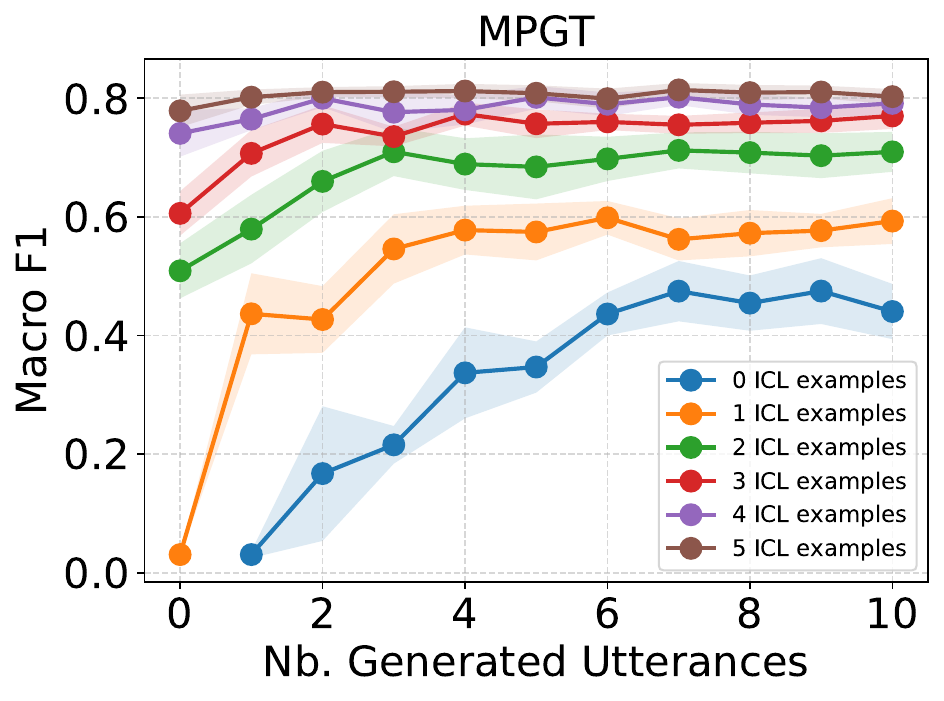}
    \end{subfigure}
        \caption{Llama-3 8B.}
    
    \end{subfigure}
    \caption{Macro-F1 score of the intent recognition models fine-tuned on the generated utterances. We evaluate different combinations on the number of augmented examples (from zero to ten) and amount of ICL examples (from zero-shot to five-shot settings). Increasing both numbers provides higher performances.}
    \label{fig:baseline_results}
    
\end{figure*}

We perform an exhaustive evaluation on the use of multiple ICL examples and generated utterances, without any disambiguation strategy. In particular, we evaluate classification models on different combinations of ICL examples --from zero to five-- and synthetic utterances --from zero to ten. Figure~\ref{fig:baseline_results} indicates that, consistently across all datasets, higher amounts of in-context learning instances and generated utterances, provide better performances.

\section{Silhouette Coefficient Analysis}
\label{sec:app_silhouette}
\begin{figure}[t]
    \centering
    \includegraphics[width=1.00\linewidth]{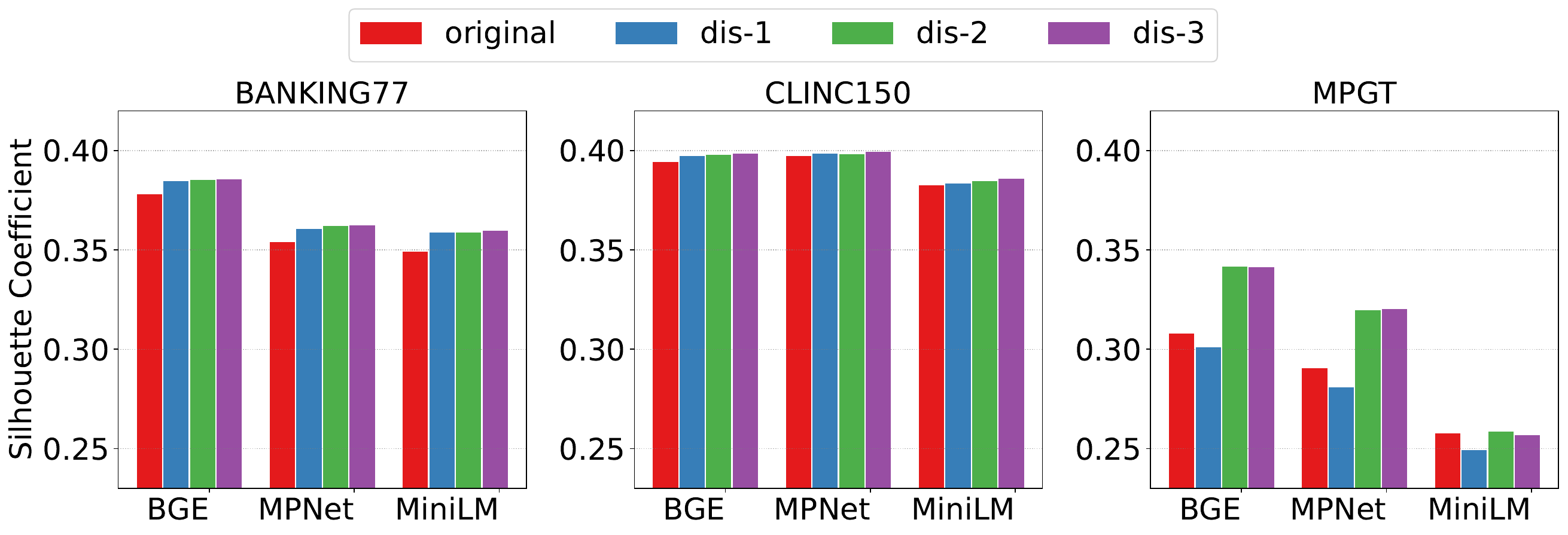}
    \caption{Silhouette coefficients of the original Mistral 7B generations and re-generations after multiple iterative disambiguation steps (dis-1, dis-2, dis-3). A higher coefficient indicates a better inter-cluster and intra-cluster mean distance relation of the utterance in the embedding space. In all scenarios, the coefficients increase after 3 disambiguation steps.}
    \label{fig:silhouette_analysis_annex}
\end{figure}

Figure~\ref{fig:silhouette_analysis_annex} shows how the silhouette coefficient varies after each disambiguation step. We note that the coefficient tends to increase as more disambiguation steps are conducted.

\section{PVI Experiments}
\label{sec:app_pvi_experiments}
\subsection{Implementation Details}
 We compare our proposed approach with the method proposed in \cite{lin-etal-2023-selective}. To do so, we conduct our own implementation of the Pointwise $\mathcal{V}$-Information (PVI) method \cite{pmlr-v162-ethayarajh22a}. PVI is a score that computes the amount of information that is carried by an instance for a classification task. In other words, the score can be used to quantify how useful a synthetic utterance is for an intent recognition task. A higher value indicates that the example provides more information, and thus that it is more useful. We use the same models described in \ref{sec:app_bert} and \ref{sec:app_llms} to compute the PVI on generated utterances. \citet{lin-etal-2023-selective} proposed two PVI methods: Global PVI, which uses a threshold based on all the examples; and Per-Intent PVI, which uses a different threshold per class. In Table~\ref{table:results_main_comb_alter}, we report the results obtained using Per-Intent PVI. We invite readers to refer to \cite{lin-etal-2023-selective,pmlr-v162-ethayarajh22a}, for more details about the PVI-based method.
 
\subsection{Results Analysis}
\begin{figure}[h!]
\centering
\includegraphics[width=0.85\linewidth]{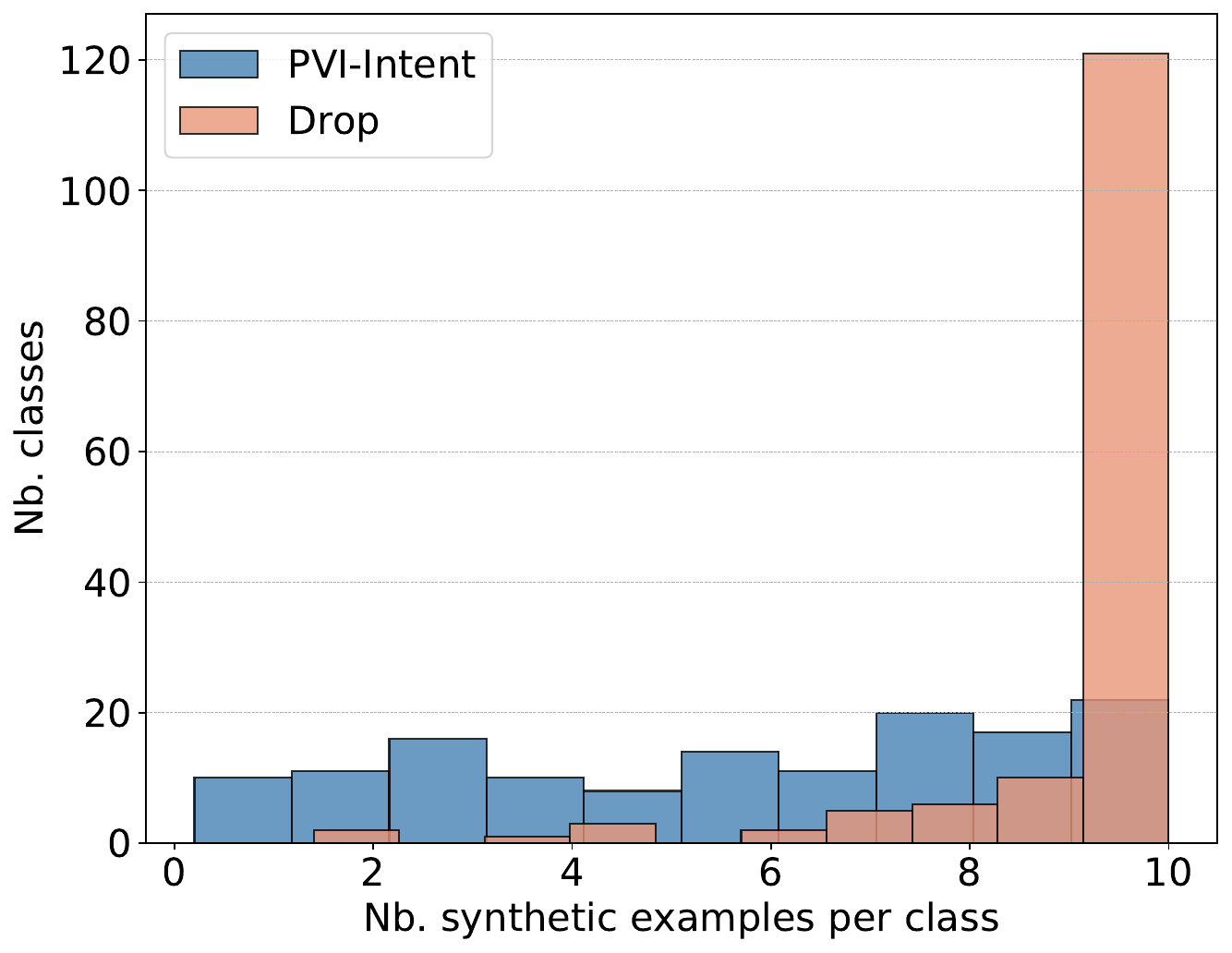}
\caption{Distribution of unfiltered examples per class after applying the PVI-Intent and Drop (with respect to BGE) methods on Mistral 7B generations over the CLINC150 corpus. The PVI-Intent method under-represents more intents than the Drop approach.}
\label{fig:class_imbalance}
\end{figure}

Intent detection models' evaluations reported in Table~\ref{table:results_main_comb_alter}, show that our approach significantly outperform our implementation of the PVI-based method. We argue that filtering out certain synthetic utterances (e.g. less informative examples), may result in problems related to class imbalance at training time. We note that, by following the PVI-based method on a given corpus, the number of examples per intent varies between 2 and 9, out of 10 originally generated utterances. Figure~\ref{fig:class_imbalance} shows the distribution of unfiltered synthetic examples per class on both the PVI-Intent and Drop methods. We observe that the PVI-Intent approach exhibits an under-representation of more classes than our Drop strategy. In other words, certain classes are relatively more represented than others at fine-tuning in the classification task. These findings are observed in all corpora, across all LLM and sentence encoder settings. Note that our re-generation strategies do not discard examples, contrary to the PVI-based filtering and Drop methods. Thus, the re-generation strategy do not induce any class imbalance.

Additionally, we believe that the computed performances are lower than the ones reported in \cite{lin-etal-2023-selective} because our experiments consider different configurations. For instance, their implementation use larger models than the ones used in this work: OPT-66B \cite{zhang2022opt} and RoBERTa\textsubscript{LARGE} \cite{liu2019roberta} as generator and classification model, respectively. We have made every effort to faithfully implement the method proposed in \citet{lin-etal-2023-selective}. However, as with any reproduction, there remains the possibility of unintentional discrepancies.

\section{MPGT Results Analysis}
\label{sec:app_ambiguity_by_nb_icl_mpgt}
We observed in Table~\ref{table:results_main_comb_alter} that the MPGT corpus shows the highest performance lifts when conducting a disambiguation strategy. One key difference between the BANKING77 or CLINC150 datasets, and MPGT, is the semantic similarity among intents. While BANKING77 and CLINC150 comprise 77 and 150 intents, respectively, the MPGT dataset includes only 8 intents. Intents in the MPGT corpus include: \texttt{affirm}, \texttt{bye thank}, \texttt{cant-help}, \texttt{incomplete-da}, \texttt{inform}, \texttt{offer-help}, \texttt{request}, \texttt{suggest}. On the other hand, BANKING77 corresponds to one single domain, and CLINC150 to 10. We argue that our method shows performance gains in scenarios with loosely or broadly defined intents.

\begin{figure}
    \centering
    \begin{subfigure}[t]{0.66\linewidth}
        \centering
        \includegraphics[width=1.0\linewidth]{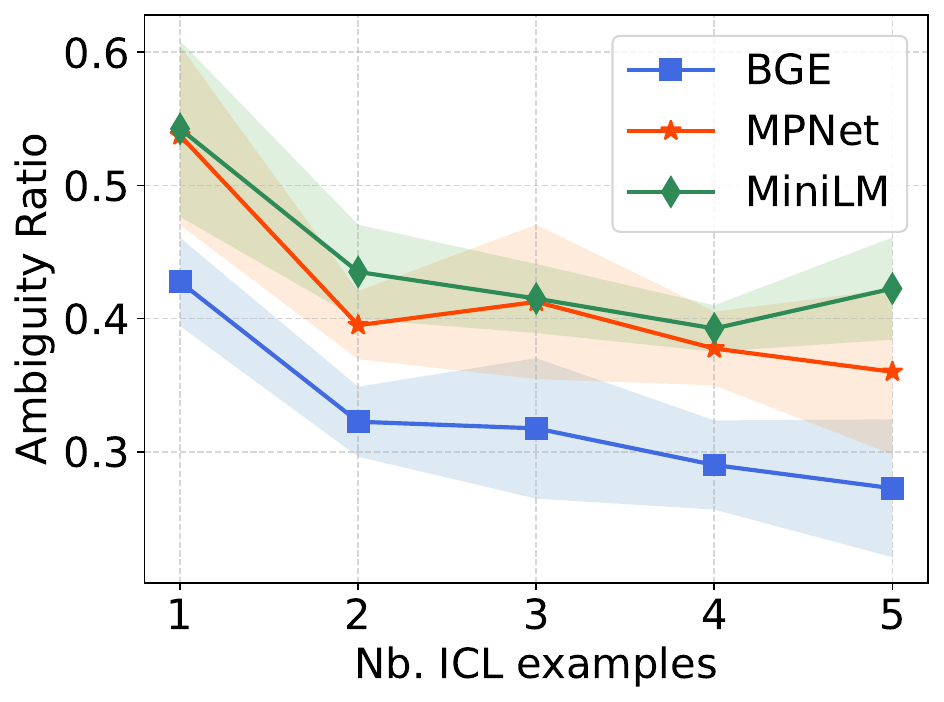}
        \caption{Mistral 7B.}
        % \label{fig:turns}
    \end{subfigure}
    \begin{subfigure}[t]{0.66\linewidth}
        \centering
        \includegraphics[width=1.0\linewidth]{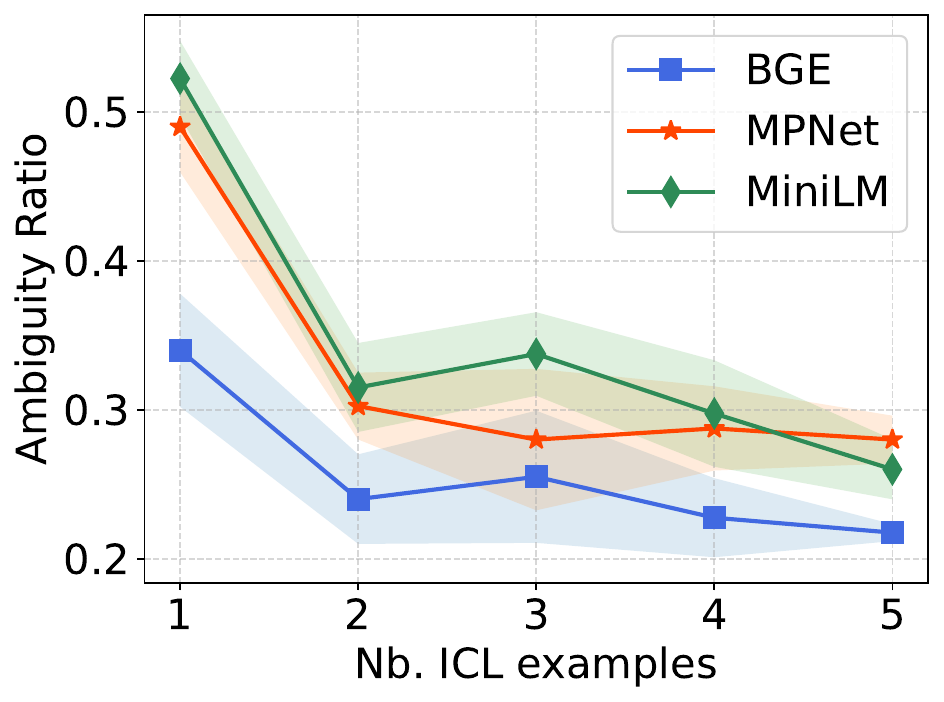}
        \caption{Llama-3 8B.}
    \end{subfigure}
\caption{Ambiguity ratio by number of in-context learning examples on LLM generations for the MPGT corpus on multiple embedding spaces. A lower ratio indicates a lower proportion of ambiguous generated utterances.}
\label{fig:ambiguity_by_nb_icl_mpgt}
\end{figure}

\begin{figure}
    \centering
    \begin{subfigure}[t]{0.75\linewidth}
        \centering
        \includegraphics[width=1.0\linewidth]{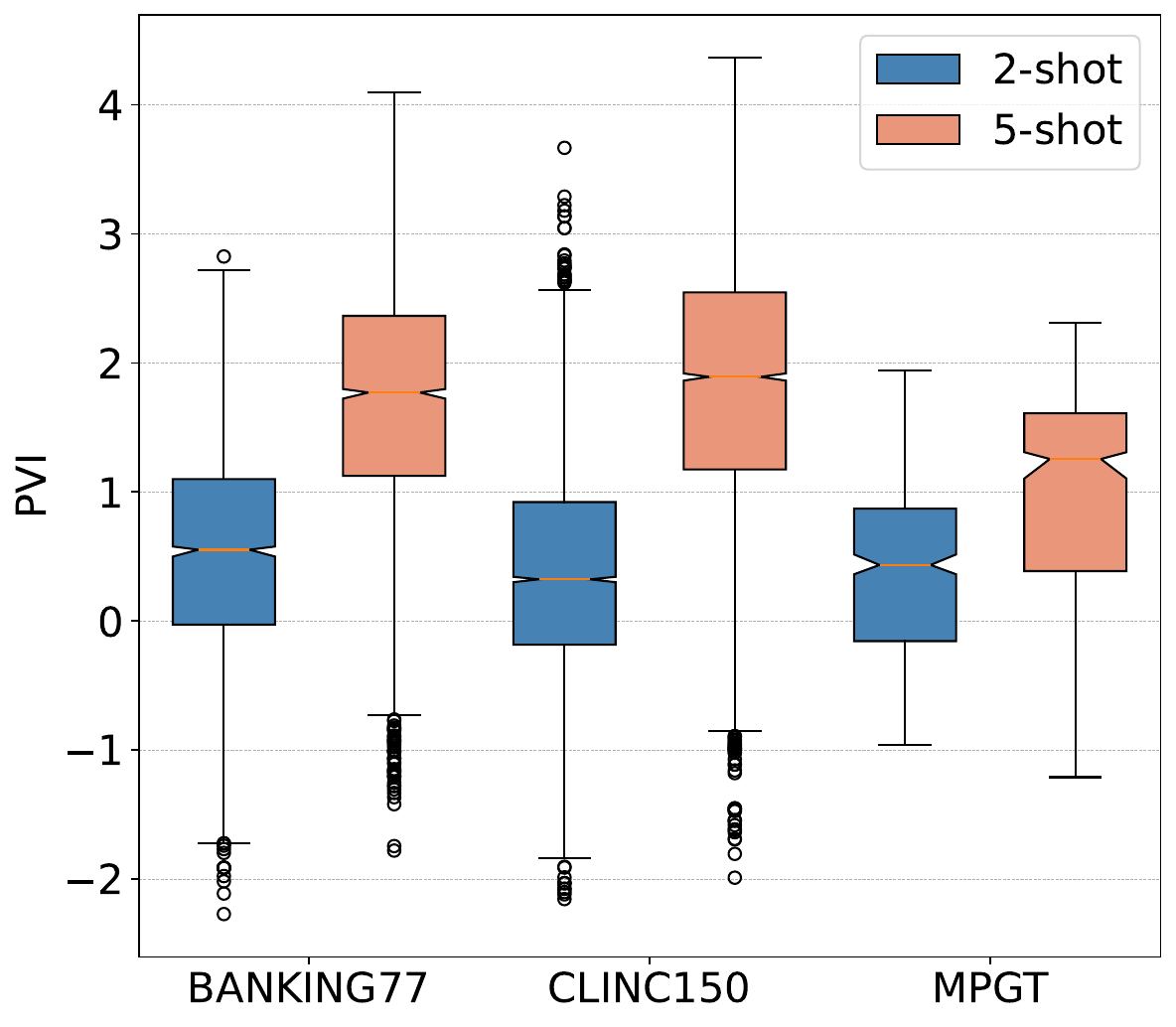}
        \caption{Mistral 7B.}
        % \label{fig:turns}
    \end{subfigure}
    \begin{subfigure}[t]{0.75\linewidth}
        \centering
        \includegraphics[width=1.0\linewidth]{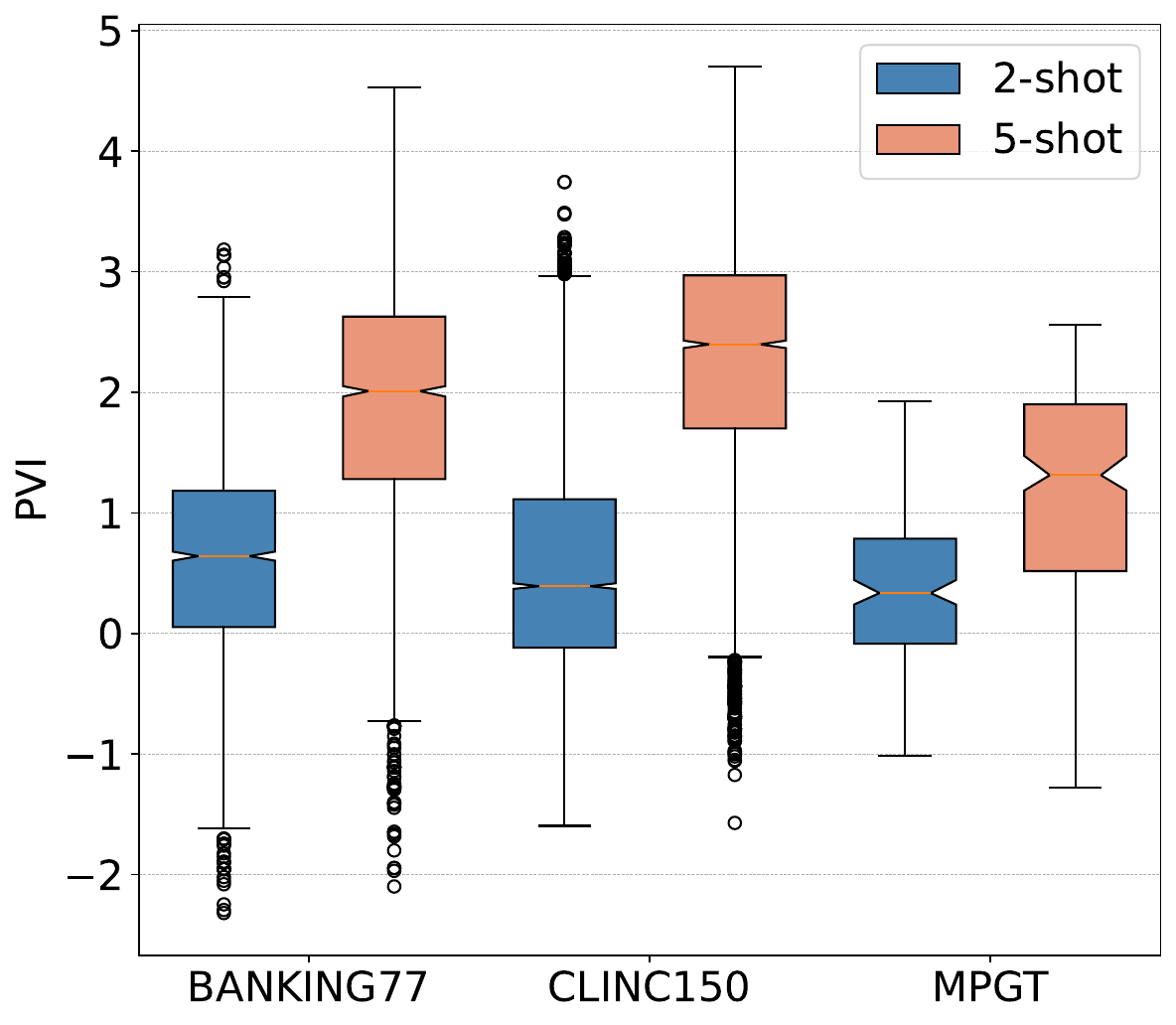}
        \caption{Llama-3 8B.}
        % \label{fig:turns}
    \end{subfigure}
\caption{PVI scores of the synthetic examples for each dataset in 2-shot and 5-shot settings. Higher values indicate that utterances are more informative (i.e., more useful) for the classification task. The MPGT corpus shows the lowest PVI values, suggesting a higher difficulty to generate good quality utterances.} 
\label{fig:pvi_corpora}
\end{figure}

We also found that the MPGT corpus presents a higher ambiguity ratio, in comparison to the other corpora. Figure~\ref{fig:ambiguity_by_nb_icl_mpgt} shows the ambiguity ratio by number of ICL examples. Following \cite{pmlr-v162-ethayarajh22a}, we compute the Pointwise $\mathcal{V}$-Information (PVI) score to asses the information carried by the synthetic examples for each corpus for our classification task. Figure~\ref{fig:pvi_corpora} indicates that the synthetic examples for the MPGT dataset present the lowest PVI scores, evidencing that generating useful utterances for the intent recognition task is more difficult on the MPGT corpus. Also, we see that the PVI scores on the 5-shot setups are much larger than in the 2-shot scenarios, which confirms that using more ICL examples lets the generators produce more useful utterances.

\section{Dataset Statistics}
\label{sec:dataset_info}
Table~\ref{table:dataset_stats} shows the number of examples per dataset split.

\begin{table}[htb]
\footnotesize
\centering
\renewcommand{\arraystretch}{1.5}  % Provide more space between table rows, if you prefer
\begin{tabular}{lrrr} \hline
% &\multicolumn{3}{c}{MIntRec2.0}&\multicolumn{3}{c}{MPGT} \\ \hline
% & \#dial & \#utt & \#utt & \%OOS \\ \hline
 & train & dev & test \\ \hline
BANKING77 & 8K & 2K & 3K \\
CLINC150 & 15K & 3K & 5K \\
MPGT & .5K & .1K & .2K  \\

\hline\end{tabular}
\caption{Number of examples per dataset split.}
\label{table:dataset_stats}
\end{table}

\section{Re-generation Qualitative Analysis}
\label{sec:qualitative_analysis}
Table~\ref{table:qualitative_analysis} shows examples of ambiguous utterances disambiguated in one and two disambiguation steps. The table includes the original utterance generation (iteration 1), the target label, the most similar label with respect to BGE, and the subsequent re-generation iterations. We observe that ambiguities occur for two main reasons: the use of ambiguous terms that may relate to unintended labels, and the inherent semantic overlap of intents in the label space.

\begin{table*}[ht!]
\tiny
\centering
\renewcommand{\arraystretch}{1.8}  % Provide more space between table rows, if you prefer
\begin{tabular}{clllllll} \hline
% &\multicolumn{3}{c}{MIntRec2.0}&\multicolumn{3}{c}{MPGT} \\ \hline
% & \#dial & \#utt & \#utt & \%OOS \\ \hline
dataset & original (it. 1) & target label & sim. label (it. 1) & re-generation (it. 2) & sim. label (it. 2) & re-generation (it. 3) & sim. label (it. 3) \\ \hline

 \multirow{3}{*}{\rotatebox[origin=c]{90}{BANKING77}} & What places accept & card & supported cards & Where can I use & card & -- & -- \\

& payments with my & acceptance & and currencies & this banking card & acceptance & & \\

& banking card? & & & for payments? & & & \\

\hline
 \multirow{3}{*}{\rotatebox[origin=c]{90}{CLINC150}} & What are the home-related & reminder & todo list & What are the home-related & reminder & -- & -- \\

& tasks that I need to & & reminder & tasks that need to be & & & \\

& remember to do? & & & reminded to be done? & & & \\

\hline

\multirow{3}{*}{\rotatebox[origin=c]{90}{CLINC150}} & What is the due & bill due & payday & What is the due & payday & What is the due & bill due  \\

& date for my next & & & date for my next & & date for my next & \\

& bank account payment? & & & bank account payment? & & bank account bill payment? & \\

\hline

\multirow{4}{*}{\rotatebox[origin=c]{90}{CLINC150}} & Can you show me my & spending & transactions & What are the details of & transactions & What are the details of & spending  \\

& spending history on & history & & my spending history on & & my spending in the & history \\

& transactions in the last & & & transactions in the last & & last month, in food and & \\

& month? & & & month in the bank? & & transportation? & \\

\hline

\multirow{2}{*}{\rotatebox[origin=c]{90}{MPGT}} & Yes, I can help & affirm & offer help & Absolutely, I can help & offer help & Absolutely & affirm  \\

& with that. & & & you with that. & & & \\

\hline\end{tabular}
\caption{Qualitative analysis of our disambiguation strategy. Original utterances and re-generations correspond to the first generations (iteration 1) and subsequent  disambiguated utterances (iterations 2 and 3) produced by Mistral 7B, respectively. The target label is included, as well as the most similar label computed after each iteration with respect to BGE.}
\label{table:qualitative_analysis}
\end{table*}

\section{Disambiguation Computational Cost}
\label{sec:computational_cost}
In this section we discuss the cost-benefit trade-off of our disambiguation strategy. While our approach increases the number of LLM calls for data augmentation, the motivation of this work is to produce higher-quality training data that enables small, efficient models --e.g., BERT-based classifiers-- to achieve substantially better intent recognition performance without relying on LLMs at inference time, as proposed in previous work \cite{arora-etal-2024-intent, castillo-lopez-etal-2025-intent,sali-toraman-2025-navigating}. Therefore, the one-time augmentation cost is amortized over the lifetime of a much more efficient deployment, an advantage aligned with prior work showing the long-term utility of high-quality synthetic data for small, or hybrid (LLMs + smaller models), intent recognition systems. 

In addition, only a subset of examples require more than one iteration, and empirically this fraction is small in corpora with large label spaces such as BANKING77 and CLINC150. Thus, the overhead is limited in practice. The computational cost due to our re-generative approach is directly proportional to the ambiguity ratio described in Figure~\ref{fig:ambiguity_ratios}. Table~\ref{table:computational_cost} shows the cumulative percentual additional costs due to the iterative procedure with respect to the original data augmentation, when using Mistral 7B and BGE on all the datasets.

\begin{table}[H]
\footnotesize
\centering
\renewcommand{\arraystretch}{1.5}  % Provide more space between table rows, if you prefer
\begin{tabular}{lrrr} \hline
%  & dis-1 & dis-2 & dis-3 \\ \hline
% BANKING77 & 13.2\% & 11.6\% & 31.3\% \\
% CLINC150 & 23.2\% & 19.9\% & 58.1\% \\
% MPGT & 32.2\% & 27.3\% & 81.7\%  \\

 & BANKING77 & CLINC150 & MPGT \\ \hline
dis-1 & 13.2\% & 11.6\% & 31.3\% \\
dis-2 & 23.2\% & 19.9\% & 58.1\% \\
dis-3 & 32.2\% & 27.3\% & 81.7\%  \\

\hline\end{tabular}
\caption{Computational costs of the disambiguation strategy with Mistral 7B and BGE on all the datasets. Values correspond to the cumulative percentual additional costs across iterations (costs are accumulated after each iteration) due to the iterative procedure \textbf{with respect to the original data augmentation}, i.e. without disambiguation.}
\label{table:computational_cost}
\end{table}

% | Disambiguation step      | BANKING77       | CLINC150      | MPGT  |
% |-------------|-------------|-------------|-----|
% | Dis-1  | 13.2% | 11.6% |  31.3% |
% | Dis-2  | 23.2% | 19.9% |  58.1% |
% | Dis-3  | 32.2% | 27.3% |  81.7% |

% \section{Example Appendix}
% \label{sec:appendix}

% This is an appendix.

\end{document}